%% file: main.tex
\documentclass[sigconf, screen]{acmart}

\acmSubmissionID{411}

\input{Preamble/packages}
\input{Preamble/definitions}

\input{Preamble/copyright}
\input{Preamble/authors}
\input{Preamble/meta}

\title[An Empirical Evaluation of Cross-City POI Recommendation]{\texorpdfstring{An Empirical Evaluation of Cross-City POI\\ Recommendation on a Large-Scale Benchmark}{An Empirical Evaluation of Cross-City POI Recommendation on a Large-Scale Benchmark}}

\begin{document}

\begin{abstract}

Cross-city \ac{POI} recommendation is crucial for navigating unfamiliar urban environments, yet its progress has historically been constrained by data limitations. Using the recently proposed large-scale benchmark \textsc{Trip World}~\cite{tripworld}, we empirically re-examine whether conclusions drawn on small prior benchmarks still hold under worldwide coverage, low home--destination region overlap, and large, semantically rich POI inventories. Our evaluation surfaces three bottlenecks of representative state-of-the-art methods: (1) hometown-aware models appear to rely more on destination-region priors than on user-specific preference transfer; (2) their accuracy--efficiency trade-off degrades at this scale, where the simplest model is among the strongest; and (3) existing mechanisms for integrating semantic metadata yield little benefit. We further include a diagnostic pilot on agentic methods adapted from next-\ac{POI} recommendation, finding that naive adaptation trails a simple popularity prior even though the relevant semantic signal is present in the data. These results highlight the need for task-specific designs that support cross-city preference transfer, semantic grounding, and scalable reasoning over unseen destination inventories.

\end{abstract}

\maketitle

\acresetall

\input{Sections/01-introduction}
\input{Sections/02-related}
\input{Sections/06-experiment}
\input{Sections/07-conclusion}

% \begin{acks}

% \end{acks}

\clearpage
\bibliographystyle{ACM-Reference-Format}
\balance
\bibliography{references}

\end{document}

%% file: Preamble/packages.tex
\usepackage{array}
\usepackage{graphicx} % Required for including images
\usepackage[skip=1pt]{caption}
\usepackage{subcaption} % Required for creating subfigures
\usepackage[inline]{enumitem}

\usepackage{bbm}
\usepackage{nameref,hyperref}
\PassOptionsToPackage{table}{xcolor}

\usepackage[normalem]{ulem}

\usepackage{csquotes}

\usepackage{listings}

\usepackage{acronym}
\usepackage{tabularx}
\usepackage{longtable}
\usepackage{tcolorbox}
\usepackage{xcolor}
\usepackage{siunitx}
\usepackage{pifont}

%% file: Preamble/definitions.tex
\AtBeginDocument{%
  \providecommand\BibTeX{{%
    \normalfont B\kern-0.5em{\scshape i\kern-0.25em b}\kern-0.8em\TeX}}}

\acrodef{GNN}{graph neural network}
\acrodef{LBSN}{location-based social network}
\acrodef{LLM}{large language model}
\acrodef{POI}{point-of-interest}
\acrodef{RNN}{recurrent neural network}
\acrodef{PEFT}{parameter-efficient fine-tuning}
\acrodef{NF}{NormalFloat}
\acrodef{FP}{floating point}
\acrodef{BF}{BrainFloating}
\acrodef{LoRA}{low-rank adaptation}
\acrodef{RFT}{reinforcement fine-tuning}
\acrodef{SFT}{supervised fine-tuning}
\acrodef{RL}{reinforcement learning}
\acrodef{MRR}{mean reciprocal rank}
\acrodef{RR}{reciprocal rank}
\acrodef{SID}{semantic ID}
\acrodef{SOM}{self organizing map}
\acrodef{RSOM}{residual self organizing map}
\acrodef{NICC}{normalized intra-class compactness}
\acrodef{NICS}{normalized inter-class separation}
\acrodef{NLL}{Negative Log-Likelihood}
\acrodef{OOT}{out-of-town}
\newcommand{\headernodot}[1]{\vspace*{1.5mm}\noindent\textbf{#1}}
\newcommand{\header}[1]{\headernodot{#1.}}

%% file: Preamble/copyright.tex
\setcopyright{rightsretained}
\copyrightyear{2026}
\acmYear{2026}
\acmConference[SIGSPATIAL '26]{The 34th ACM International Conference on Advances in Geographic Information Systems}{November 3--6, 2026}{Riverside, CA, USA}
\acmBooktitle{The 34th ACM International Conference on Advances in Geographic Information Systems (SIGSPATIAL '26), November 3--6, 2026, Riverside, CA, USA}
\acmISBN{}
\acmDOI{}

%% file: Preamble/authors.tex
\author{Peibo Li}
\orcid{0009-0001-0201-3564} 
\affiliation{%
  \institution{University of New South Wales}
  \city{Sydney}
  \country{Australia}
}
\email{peibo.li@student.unsw.edu.au}

\author{Yang Song}
\orcid{0000-0003-1283-1672}
\affiliation{%
  \institution{University of New South Wales}
  \city{Sydney}
  \country{Australia}
}
\email{yang.song1@unsw.edu.au}

\author{Hao Xue}
\orcid{0000-0003-1700-9215} 
\affiliation{%
  \institution{The Hong Kong University of Science and Technology (Guangzhou)}
  \city{Guangzhou}
  \country{China}
}
\email{haoxue@hkust-gz.edu.cn}

\author{Maarten de Rijke}
\orcid{0000-0002-1086-0202} 
\affiliation{%
  \institution{University of Amsterdam}
  \city{}
  \country{The Netherlands}
}
\email{m.derijke@uva.nl}

\author{Flora D. Salim}
\orcid{0000-0002-1237-1664}
\affiliation{%
  \institution{University of New South Wales}
  \city{Sydney}
  \country{Australia}
}
\email{flora.salim@unsw.edu.au}

\renewcommand{\shortauthors}{Peibo Li et al.}

%% file: Preamble/meta.tex
\ccsdesc[500]{Information systems~Recommender systems}

\keywords{Point-of-interest recommendation}

%% file: Sections/01-introduction.tex
\section{Introduction}
\label{Section:intro}
Point-of-interest (POI) recommendation underpins location-based services, helping users discover relevant places~\cite{zhao2016survey,zhang2025survey}. Most work targets the in-city or next-POI setting, where users repeatedly interact with POIs in their home region~\cite{wang2017location,li2020deep}. Many real travel scenarios break this assumption: a traveler arriving in an unfamiliar city has no prior interactions with the destination POIs, only a short and sparse trajectory, and preferences that shift from everyday routines to tourism. Cross-city POI recommendation therefore requires transferring user preferences from a home city to a destination with a largely unseen POI inventory~\cite{ding2019learning,li2020deep,sun2024city}, while contending with habitual histories that do not translate to travel behavior, geographically and semantically disjoint POIs, and limited sequential evidence~\cite{xin2022captor,liu2025spot}.

Progress on this setting has been limited by the lack of large-scale, realistic, and semantically rich benchmarks. Existing datasets are designed for single-city or in-city next-POI prediction~\cite{chen2016learning,thomee2016yfcc100m}, or are small, geographically narrow, and metadata-poor~\cite{yang2015nationtelescope,yang2016participatory,wang2017location}. Some provide only coordinates and coarse categories, while others offer rich reviews but rely on review-derived interactions that need not correspond to real visit sequences. These limitations make it hard to evaluate whether a method generalizes across cities, exploits semantic POI information, or handles the extreme cold start travelers face.

We conduct our evaluation on \textsc{Trip World}~\cite{tripworld}, a recently proposed large-scale benchmark that provides worldwide out-of-town check-ins together with enriched POI metadata and user reviews, to rigorously study cross-city POI recommendation. Its scale and coverage support a unified comparison of classical, neural, and recent agentic methods, and enable testing whether models leverage mobility history, destination context, POI semantics, and travel-specific behavior beyond simple proximity or popularity signals. Our contributions are:
\begin{itemize}
    \item We evaluate representative cross-city POI recommendation methods on \textsc{Trip World}, spanning popularity-based, sequential, neural, semantic-aware, and \ac{LLM}-agentic approaches.
    \item We surface three bottlenecks: hometown-aware methods appear to rely more on destination-region priors than on transferable user preferences; the simplest model is among the strongest while the most compositional ranks lowest at much higher cost; and existing semantic-integration mechanisms yield little benefit.
    \item As a diagnostic pilot, we benchmark agentic methods adapted from next-POI recommendation and show that naive transfer is insufficient, charting directions for cross-city-native designs.
\end{itemize}

%% file: Sections/02-related.tex
\section{Related Work}

\subsection{Trip and Cross-City POI Recommendation}
Intra-city trip recommendation predicts a sequence of intermediate POIs given an origin, destination, and budget within one region. Learning-based methods evolved from combinatorial optimization to deep sequence models: MatTrip~\cite{zhang2024encoder} (a dual-LSTM encoder-decoder), Graph-Trip~\cite{gao2023dual} (heterogeneous POI graphs), AR-Trip~\cite{shu2024analyzing} (a Transformer mitigating repetition), and PPROC~\cite{iakovlev2025learning} (neural spatiotemporal point processes). All rely on rich localized history and degrade under the cold start of out-of-town travel.

Cross-city (out-of-town) recommenders transfer preferences from a data-rich hometown to an unfamiliar target city across disjoint POI spaces. Early generative and content-aware models share a city-agnostic latent space~\cite{yin2014joint,gao2015content,yin2017spatial}; graph and causal approaches disentangle hometown preference from destination constraints~\cite{xin2021out,liu2024kddc}, and meta-learning transfers preference initializations across cities~\cite{wang2019cross,chen2021curriculum,ding2022cross}. Most target isolated next-POI prediction; SPOT-Trip~\cite{liu2025spot} instead unifies trip modeling with cross-city transfer via a POI attribute knowledge graph and neural ODEs for preference drift, reaching state of the art in data-scarce regions.

\subsection{Datasets for Trip and Cross-City Recommendation}
Table~\ref{tab:datasets} summarizes six widely used datasets. Edinburgh, Glasgow, Osaka, and Toronto are reconstructed from Flickr photos~\cite{thomee2016yfcc100m,chen2016learning} and each cover a single region with only hometown check-ins. Only Foursquare~\cite{yang2015nationtelescope,yang2016participatory} and Yelp\footnote{\url{https://www.yelp.com/dataset}} contain genuine cross-city activity, so we adopt them for comparison; both are small and largely US-centric. Foursquare carries no auxiliary \ac{POI} information, whereas Yelp offers rich reviews but its check-ins are review artifacts rather than real visits, with roughly 50\% of out-of-town behavior being ``phantom travel'' within the same metropolitan area. These gaps motivate a large-scale, worldwide, and semantically rich benchmark such as \textsc{Trip World}~\cite{tripworld}, which we adopt for our evaluation.

\begin{table}[t]
    \centering
    \small
    \caption{Statistics of prior datasets. CIs denote check-ins; Home and \ac{OOT} are hometown and out-of-town check-ins. Real CI indicates whether the dataset contains genuine check-in events (\ding{51}) versus reconstructed or proxy records (\ding{55}). The four city-specific datasets cover a single region and contain only hometown check-ins by construction.}
    \label{tab:datasets}
    \setlength{\tabcolsep}{3pt}
    \begin{tabular}{lrrrrrrc}
        \toprule
        Dataset & Users & Regions & POIs & CIs & Home & OOT & Real CI \\
        \midrule
        Edinburgh   & $1{,}454$ & $1$   & $28$       & $33{,}944$  & $33{,}944$  & $0$        & \ding{55} \\
        Glasgow     & $601$     & $1$   & $27$       & $11{,}434$  & $11{,}434$  & $0$        & \ding{55} \\
        Osaka       & $450$     & $1$   & $27$       & $7{,}747$   & $7{,}747$   & $0$        & \ding{55} \\
        Toronto     & $1{,}395$ & $1$   & $29$       & $39{,}419$  & $39{,}419$  & $0$        & \ding{55} \\
        Foursquare  & $3{,}007$ & $21$  & $23{,}884$ & $126{,}219$ & $109{,}225$ & $16{,}994$ & \ding{51} \\
        Yelp        & $4{,}417$ & $214$ & $29{,}930$ & $78{,}882$  & $58{,}403$  & $20{,}479$ & \ding{55} \\
        \bottomrule
    \end{tabular}
\end{table}

%% file: Sections/06-experiment.tex
\section{Experiments}
\begin{table}[t]                                                                                                                                                       
  \centering                                                                                                                                                         
  \caption{Statistics of \textsc{Trip World} after processing. CIs: check-ins; Avg $|c_o|$: mean number of POIs per out-of-town trip.}
  \label{tab:crosscityclean_stats}                                                                                                                                        
  \setlength{\tabcolsep}{1pt}                                                                                                                                              
                                                                                                                                            
  \begin{tabular}{lcccccc}                                                                                                                                           
  \toprule                                                                                                                                                                 
  Dataset & \#Users & \#POIs & \#Regions   & \#Home CIs & \#Travel CIs & Avg \(|c_o|\)  \\
  \midrule                                                                                                                                                                 
  \textsc{Trip World} & 63{,}896 & 336{,}102 & 890 &  6{,}242{,}544 & 576{,}504 & 4.97  \\                                                                                                                                                             
  \bottomrule                                                                                                                                                              
  \end{tabular}%                                                                                                                                                                                                                                                                                          
  \end{table}    
\subsection{Experimental setup}
\subsubsection{Task}
Given a user's hometown check-in history $H_u$ and an out-of-town trip specified by its origin $o$, destination $d$, target region $r$, and length $L$, the task is to predict the ordered intermediate POIs $c_o[1{:}{-}1]$ visited between $o$ and $d$. All models observe $H_u$, $o$, $d$, $r$, and $L$; semantic-aware variants additionally observe POI categories, coordinates, and review-derived signals. Evaluation is over the intermediate POIs only.
\subsubsection{Baselines}
We benchmark 7 baselines: 4 trip recommendation methods without hometown information (Popularity~\cite{chen2016learning}, GraphTrip~\cite{gao2023dual}, MatTrip~\cite{zhang2024encoder}, AR-Trip~\cite{shu2024analyzing}) and 3 with hometown information (KDDC~\cite{liu2024kddc}, PPROC~\cite{iakovlev2024learning}, SPOT-Trip~\cite{liu2025spot}), where KDDC and PPROC use the SPOT-Trip backbone. The \textbf{Popularity} baseline ranks POIs by their global training-set visit frequency. All experiments run on a cluster of H200 GPUs.
\subsubsection{Evaluation metric}
We adopt F1 and Pairs-F1, following existing work~\cite{gao2023dual, liu2025spot, shu2024analyzing, kuo2023bert}: F1 measures the overlap between predicted and ground-truth intermediate POIs, while Pairs-F1 additionally rewards correct visiting order. Both are computed over the intermediate POIs only, excluding the fixed origin and destination.
\subsubsection{Dataset processing}
We follow~\cite{liu2025spot} to process the dataset. We filter out users with fewer than three out-of-town check-in records, and exclude out-of-town trips whose travel durations are shorter than 1 hour or longer than 30 days. We partition by user into training, validation, and testing sets with an $80\%/10\%/10\%$ ratio. For baselines that require knowledge graphs, we generate entity-specific relations following the original papers, using auxiliary information from the dataset. Statistics after processing are shown in Table~\ref{tab:crosscityclean_stats}.

\subsection{Results and Discussion}
 \begin{table*}[t]
  \centering
  \caption{Out-of-town trip recommendation performance, measured by F1 and Pairs-F1 over the intermediate POIs (higher is better, $\uparrow$). Foursquare and Yelp numbers are reproduced from Liu et al.~\cite{liu2025spot} (single published values), \emph{not} re-run in our pipeline; Trip World numbers are averaged over 3 random seeds (per-seed std ${\le}0.003$ for F1 and ${\le}0.0013$ for Pairs-F1, so the neural baselines differ by within roughly one standard deviation). \textbf{Bold} marks the best value per column. \textit{SPOT-Trip no-review} is a Trip-World-specific ablation that removes the review-rating relation and has no counterpart in the original benchmark (\textemdash).}
  \label{tab:clean_general}
  \setlength{\tabcolsep}{6pt}
  \begin{tabular}{lcccccc}
  \toprule
   & \multicolumn{2}{c}{Foursquare} & \multicolumn{2}{c}{Yelp} & \multicolumn{2}{c}{Trip World} \\
  \cmidrule(lr){2-3}\cmidrule(lr){4-5}\cmidrule(lr){6-7}
  Architecture & F1 $\uparrow$ & Pairs-F1 $\uparrow$ & F1 $\uparrow$ & Pairs-F1 $\uparrow$ & F1 $\uparrow$ & Pairs-F1 $\uparrow$ \\
  \midrule
  Popularity \cite{chen2016learning} & 0.0261 & 0.0013 & 0.0257 & 0.0056 & 0.0222{\,$\pm$\,}0.0012 & 0.0057{\,$\pm$\,}0.0005 \\
  MatTrip \cite{zhang2024encoder}    & 0.0311 & 0.0037 & 0.0301 & 0.0119 & \textbf{0.0601}{\,$\pm$\,}0.0019 & \textbf{0.0166}{\,$\pm$\,}0.0010 \\
  GraphTrip \cite{gao2023dual}       & 0.0295 & 0.0048 & 0.0289 & 0.0126 & 0.0582{\,$\pm$\,}0.0017 & 0.0164{\,$\pm$\,}0.0005 \\
  AR-Trip \cite{shu2024analyzing}    & 0.0304 & 0.0045 & 0.0307 & 0.0153 & 0.0562{\,$\pm$\,}0.0028 & 0.0151{\,$\pm$\,}0.0013 \\
  \midrule
  KDDC \cite{liu2024kddc}            & 0.0375 & 0.0079 & 0.0341 & 0.0156 & 0.0575{\,$\pm$\,}0.0005 & 0.0159{\,$\pm$\,}0.0001 \\
  PPROC \cite{iakovlev2024learning}  & 0.0330 & 0.0071 & 0.0334 & 0.0159 & 0.0554{\,$\pm$\,}0.0008 & 0.0148{\,$\pm$\,}0.0008 \\
  SPOT-Trip no-review \cite{liu2025spot} & \textemdash & \textemdash & \textemdash & \textemdash & 0.0572{\,$\pm$\,}0.0004 & 0.0159{\,$\pm$\,}0.0006 \\
  SPOT-Trip full \cite{liu2025spot}  & \textbf{0.0400} & \textbf{0.0109} & \textbf{0.0399} & \textbf{0.0190} & 0.0557{\,$\pm$\,}0.0013 & 0.0151{\,$\pm$\,}0.0010 \\
  \bottomrule
  \end{tabular}
  \end{table*}
We report baseline performance in Table~\ref{tab:clean_general}. Using the published Foursquare/Yelp results as a contextual reference rather than a controlled re-run in our pipeline, we observe that the relative ranking on Trip World differs: there, the simplest model (MatTrip) is among the strongest while the most compositional, hometown-aware SPOT-Trip ranks lowest. We examine possible explanations below.
\subsubsection{Transferable preferences, or region priors?}
On Trip World, methods that use hometown information do not outperform those relying solely on out-of-town signals, unlike on Foursquare and Yelp~\cite{liu2025spot}. The standard framing casts these methods as learning user preferences from home history and transferring them to the destination. However, as shown in Table~\ref{tab:region_overlap}, only 30.9\% of destination regions in Trip World also appear as home regions for other users, far below Foursquare and Yelp. This pattern suggests that, in this low-overlap regime, the methods rely more on region-conditional POI distributions (effectively per-region popularity priors) than on user preferences that transfer across regions. We test this directly on SPOT-Trip, the most complete hometown-aware model (KDDC and PPROC share its backbone): at test time we replace each trip's hometown history with that of a random other user, holding the trained model and decoder fixed. Across three seeds, this shuffling leaves F1 unchanged (relative change ${<}0.55\%$, within the cross-seed standard deviation), so the model is invariant to \emph{whose} home it sees and does not exploit user-specific hometown preference. We benchmark the most complete model; extending the ablation to every architecture is left to future work.
\begin{table}[t]                                                                                                                                                                                           
  \centering                                       
  \caption{Region overlap between home regions ($\mathcal{R}_h$) and destination regions ($\mathcal{R}_o$) on the three benchmarks. \emph{Dest-coverage} is the fraction of destination regions that also    
  appear as a home region. \emph{Dest-only regions} count regions that appear only as destinations and never as homes.}                                                                                     
  \label{tab:region_overlap}                                                                                                                                                                                 
  \begin{tabular}{lcc}                                                                                                                                                                                       
  \toprule                                                                                                                                                                                                   
  Dataset    & Dest-coverage $|\mathcal{R}_h \cap \mathcal{R}_o| / |\mathcal{R}_o|$ & Dest-only regions \\
  \midrule                                                                                                                                                                                                   
  Foursquare & 95.2\%  & 1   \\
  Yelp       & 69.9\%  & 56  \\                                                                                                                                                                              
  Trip World       & 30.9\%  & 603 \\                                                                                                                                                                     
  \bottomrule
  \end{tabular}                                                                                                                                                                                              
  \end{table}    

\subsubsection{Accuracy--efficiency trade-off at scale.} Trip World is an order of magnitude larger than previous benchmarks, raising the question of how the baselines behave at this scale. MatTrip, the simplest model (a Bi-LSTM with Bahdanau attention), attains the best accuracy, while the most compositional model, SPOT-Trip (a knowledge graph with an ODE decoder), ranks lowest while being the costliest to train (Table~\ref{tab:efficiency}). This is consistent with the additional inductive biases adding variance rather than reducing it in a regime where the per-POI gradient signal is sparse ($336{,}102$ POIs across $890$ regions). We cannot fully separate architecture from tuning or implementation here, and leave a controlled scaling study (e.g., training-size sweeps) to future work; nonetheless, the architectures that pay the most overhead also score lowest.

  \begin{table}[t]                                                                                                                                                                                           
  \centering                                                                                                                                                                                                 
  \caption{Per-epoch training time on Trip World (single H200 GPU).}                                                                                                                                                                                      
  \label{tab:efficiency}                                                                                                                                                                                     
  \begin{tabular}{lcc}                                                                                                                                                                                       
  \toprule                                                                                                                                                                                                   
  Architecture & s / epoch & Relative to MatTrip \\                                                                                                                                                          
  \midrule                                                                                                                                                                                                   
  MatTrip            &  58 & 1.0$\times$ \\                                                                                                                                                                  
  GraphTrip       & 178 & 3.1$\times$ \\                                                                                                                                                                  
  KDDC        & 137 & 2.4$\times$ \\                                                                                                                                                                  
  AR-Trip            & 304 & 5.2$\times$ \\                                                                                                                                                                  
  PPROC       & 330 & 5.7$\times$ \\                                                                                                                                                                                                                                                                                                       
  SPOT-Trip full     & 422 & 7.3$\times$ \\                                                                                                                                                                  
  \bottomrule                                                                                                                                                                                                
  \end{tabular}                                                                                                                                                                                              
  \end{table}   

\subsubsection{Integrating semantic metadata}
Trip World provides rich semantic features. We probe one common integration mechanism: \textit{SPOT-Trip full} adds a knowledge-graph relation encoding each POI's average rating, while \textit{SPOT-Trip no-review} omits it. Counterintuitively, \textit{SPOT-Trip no-review} outperforms \textit{SPOT-Trip full}, indicating that this particular mechanism, a single review-derived rating relation, does not help. We stress that this probes \emph{one} mechanism rather than semantic metadata in general; the richer review text and user-taste signals are examined in our agentic pilot (Section~\ref{sec:agentic}). Existing methods were not designed for such features, as they were unavailable in prior benchmarks, and we hope Trip World's semantics motivate models that can harness them.

\subsection{Can Agentic Methods Transfer to Cross-City Recommendation?}
\label{sec:agentic}
As a diagnostic pilot, we test whether \ac{LLM}-based \emph{agentic} methods, which reason over textual context and natively consume semantic metadata, alleviate the bottlenecks above. We adapt two representative methods from next-\ac{POI} prediction: LLMMove~\cite{feng2024move}, a direct \ac{LLM} ranking prompt, and AgentMove~\cite{feng2025agentmove}, a full agent with memory and world-model reasoning, both backed by Claude Haiku~4.5 (\texttt{claude-haiku-4-5}).

\header{Adaptation and setup}
Open-vocabulary generation over the $336{,}102$-POI inventory is infeasible for an \ac{LLM}, a first finding in itself. We thus recast the task as \emph{in-candidate} selection: each query receives a menu of 100 candidates guaranteed to contain the ground-truth intermediates, which the agent selects and orders. Negatives are sampled to \emph{approximately} match ground-truth popularity (residual GT-vs-negative popularity AUC ${\approx}0.55$), so popularity cannot trivially separate the answer; the \emph{in-candidate Popularity} baseline ranks candidates by leave-one-out training frequency. Because this restriction raises the chance level, absolute F1 here is \emph{not} comparable to Table~\ref{tab:clean_general}; the fair references are the in-candidate Popularity and Random baselines. Beyond trajectory context, we supply two review-grounded signals. The \emph{review summary} is a short \ac{LLM}-written description of what a POI is known for and whom it suits, distilled from its reviews and attached to every candidate. The \emph{user-taste profile} is a concise \ac{LLM}-written description of a user's preferred cuisines, venue types, vibe, and price level, distilled from the review summaries of the POIs the user visited at home, and prepended to the prompt.

\begin{table}[t]
  \centering
  \caption{Agentic methods on cross-city POI recommendation (\emph{in-candidate} setting, data-rich destinations, Claude Haiku~4.5). Values are not comparable to Table~\ref{tab:clean_general}; the references are the in-candidate Popularity and Random baselines. \textbf{Bold} marks the best value per column.}
  \label{tab:agentic_main}
  \setlength{\tabcolsep}{6pt}
  \begin{tabular}{lcc}
  \toprule
  Method & F1 $\uparrow$ & Pairs-F1 $\uparrow$ \\
  \midrule
  Random (in-candidate)            & 0.0361 & 0.0040 \\
  AgentMove~\cite{feng2025agentmove}, no semantics & 0.0627 & 0.0102 \\
  AgentMove~\cite{feng2025agentmove}, +\,semantics & 0.0658 & 0.0116 \\
  LLMMove~\cite{feng2024move}, +\,semantics         & 0.0718 & 0.0146 \\
  LLMMove~\cite{feng2024move}, +\,popularity        & 0.0820 & 0.0189 \\
  \midrule
  Popularity (in-candidate)        & \textbf{0.1069} & \textbf{0.0234} \\
  \bottomrule
  \end{tabular}
  \end{table}

\header{Naive transfer underperforms a popularity prior}
Table~\ref{tab:agentic_main} reports the in-candidate results on data-rich destinations. Every agentic variant trails the simple in-candidate Popularity prior by a wide margin (e.g., $0.0718$ vs.\ $0.1069$ F1). Echoing the scaling result for neural baselines, the more elaborate AgentMove underperforms the lighter LLMMove prompt, indicating that additional agentic machinery again adds variance rather than signal. Strikingly, injecting the candidates' popularity into the prompt (LLMMove,~+\,popularity) helps more than any semantic signal, yet still does not close the gap to the popularity baseline. Naive transfer of agentic methods is therefore \emph{insufficient} for cross-city POI recommendation.

\begin{table}[t]
  \centering
  \caption{Disentangling the two semantic signals on \emph{cold-start} destinations ($\le 30$ training trips), LLMMove with Claude Haiku~4.5 (in-candidate). Removing either signal \emph{improves} F1; the variant with no semantics is best among the \acp{LLM}, and all trail the Popularity prior.}
  \label{tab:agentic_coldstart}
  \setlength{\tabcolsep}{5pt}
  \begin{tabular}{lcc}
  \toprule
  Variant & F1 $\uparrow$ & Pairs-F1 $\uparrow$ \\
  \midrule
  Random (in-candidate)        & 0.0324 & 0.0089 \\
  \;+\,taste \;+\,review       & 0.0490 & 0.0083 \\
  \;+\,taste \;$-$review       & 0.0536 & 0.0097 \\
  \;$-$taste \;+\,review       & 0.0524 & 0.0125 \\
  \;$-$taste \;$-$review       & 0.0591 & 0.0144 \\
  \midrule
  Popularity (in-candidate)    & \textbf{0.0686} & \textbf{0.0176} \\
  \bottomrule
  \end{tabular}
  \end{table}

\header{Semantics are not exploited, but the signal is present}
To probe the semantic bottleneck, we disentangle the user-taste and review signals on the cold-start slice (destinations with $\le 30$ training trips), where transfer matters most. Table~\ref{tab:agentic_coldstart} shows that \emph{both} signals independently hurt: removing either improves F1, and the variant with no semantics is the strongest \ac{LLM} configuration. This is not a data-quality artifact, cold-start POI summaries are in fact well sourced (a median of 342 reviews per summarized POI). Rather, the agent cannot convert the metadata into correct selections and is instead distracted away from the popularity prior. Crucially, the information is \emph{present}: the ground-truth intermediates are measurably aligned with the user's hometown taste, matching the user's home POI categories $40.6\%$ of the time versus $25.3\%$ for the popularity-matched negatives. The limitation is therefore one of \emph{method}, not of available signal.

\header{Where the headroom lies}
The dominance of popularity is itself partly an artifact of the benchmark and metric. Popularity still wins overall only because the candidate matching is approximate and most queries have popular ground truth: stratifying by ground-truth popularity, the prior is strong where the answer is itself popular, but on the long-tail third of queries it falls to near the Random floor, where the \ac{LLM} agents \emph{exceed} it. This is where semantic reasoning should help, pointing to directions for cross-city-native methods: (i)~anchor on the popularity prior and use semantics only to \emph{refine} among comparable candidates, rather than letting free-form generation override a reliable prior; (ii)~align POI semantics to the user-taste vocabulary, since current review summaries describe service sentiment largely orthogonal to the preferences that distinguish the ground truth; and (iii)~develop scalable retrieval and grounding to relax the in-candidate restriction toward open-vocabulary cross-city recommendation. These mirror the three bottlenecks found for neural baselines, preference transfer, semantic grounding, and scale.

%% file: Sections/07-conclusion.tex
\section{Conclusion}
Using \textsc{Trip World}, a large-scale, worldwide, and semantically rich benchmark, we re-examined state-of-the-art methods for cross-city POI recommendation. Our evaluation surfaces three bottlenecks. First, under \textsc{Trip World}'s low home--destination region overlap, hometown-aware methods rely on per-region popularity priors rather than transferable user preferences: replacing a user's hometown history with a random other user's leaves SPOT-Trip---the most complete such model---essentially unchanged, confirming directly that it does not exploit user-specific home signal. Second, their accuracy--efficiency trade-off degrades at this scale: the simplest model (MatTrip) is among the strongest, while the most compositional (SPOT-Trip) ranks lowest at the highest cost. Third, a common mechanism for integrating semantic metadata yields no benefit. A diagnostic pilot on agentic \ac{LLM} methods fares no better: they trail a simple in-candidate popularity prior, even though the relevant semantic signal is present in the data. Overall, progress will require task-specific designs for cross-city preference transfer, semantic grounding, and scalable reasoning over unseen destination inventories.